%% file: samplepaper.tex
\documentclass[runningheads]{llncs}
\usepackage[T1]{fontenc}
\usepackage{appendix}
\usepackage{graphicx}
\usepackage[ruled,vlined,linesnumbered]{algorithm2e}
\usepackage{float}
\usepackage{amsmath} 
\usepackage{amssymb} 
\usepackage{algorithmic}
\usepackage{makecell}
\usepackage{booktabs}

\begin{document}
\title{Feature-Selective Representation Misdirection for Machine Unlearning}
%
%

\author{Taozhao Chen\inst{1} \and
Linghan Huang\inst{1} \and Kim-Kwang Raymond Choo\inst{2}\and
Huaming Chen\inst{1}}
\institute{
University of Sydney, Sydney, NSW, Australia\\
\email{\{tche8294, lhua5130\}@uni.sydney.edu.au}, \email{huaming.chen@sydney.edu.au}
\and
University of Texas at San Antonio, San Antonio, TX, USA\\
\email{raymond.choo@fulbrightmail.org}
}

\maketitle              
\begin{abstract}
\input{section/abstract}

\end{abstract}

\section{Introduction}

\input{section/introduction}

\section{Related Work}

\input{section/background}

\begin{figure}[H]
    \centering
    \includegraphics[width=1\linewidth]{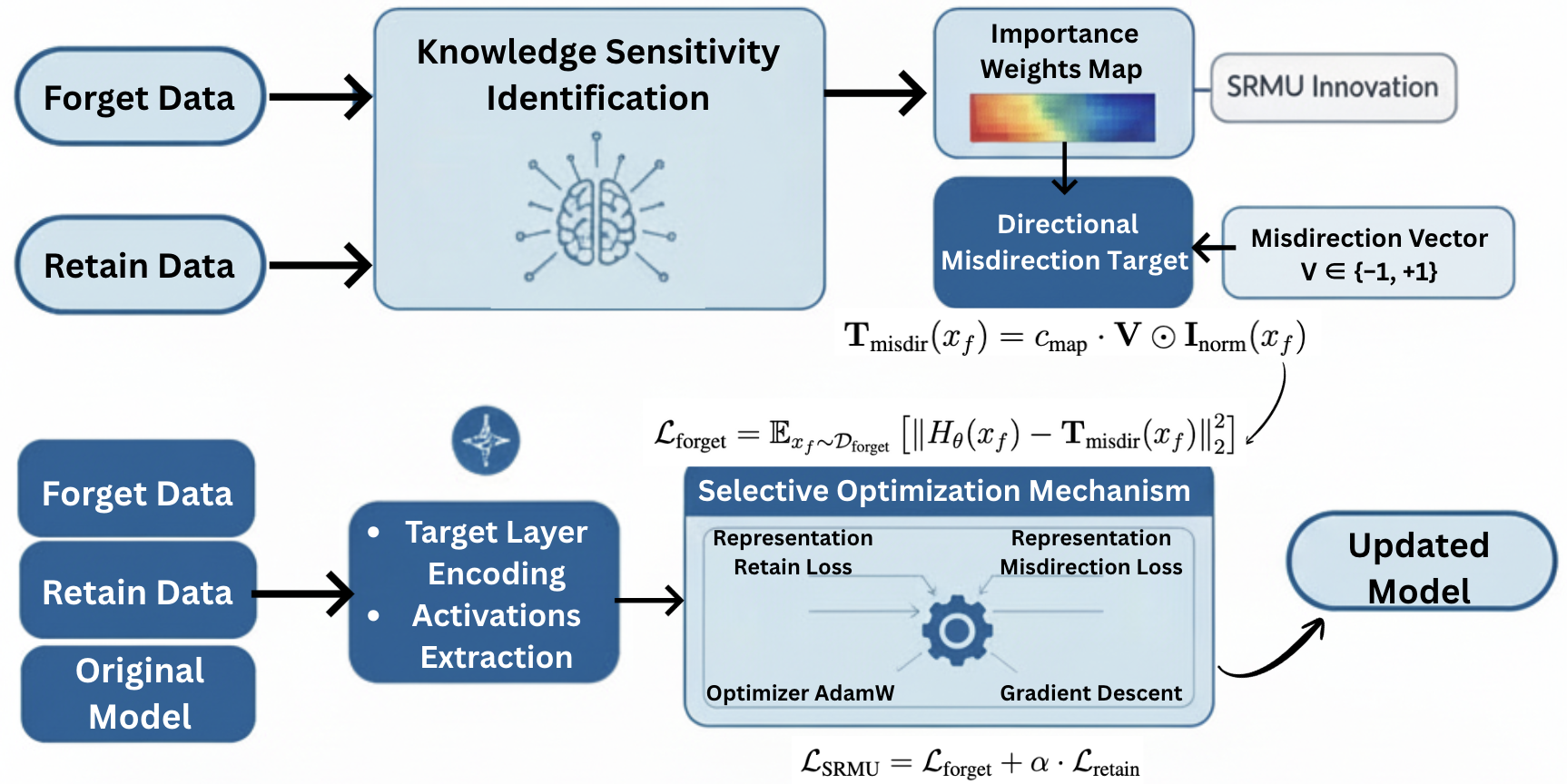}
    \caption{Overview of the proposed SRMU framework for target unlearning.}
    \label{fig:selectivermu}
\end{figure}
\section{Methodology}

\input{section/method}

\section{Evaluation}
\input{section/new_experiment}

\section{Conclusion}
\input{section/conclusion}

\bibliographystyle{splncs04} 
\bibliography{ref}

\end{document}

%% file: section/abstract.tex

As large language models (LLMs) are increasingly adopted in safety-critical and regulated sectors, the retention of sensitive or prohibited knowledge introduces escalating risks, ranging from privacy leakage to regulatory non-compliance to to potential misuse, and so on. Recent studies suggest that machine unlearning can help ensure deployed models comply with evolving legal, safety, and governance requirements. However, current unlearning techniques assume clean separation between forget and retain datasets, which is challenging in operational settings characterized by highly entangled distributions. In such scenarios, perturbation-based methods often degrade general model utility or fail to ensure safety. To address this, we propose Selective Representation Misdirection for Unlearning (SRMU), a novel principled activation-editing framework that enforces feature-aware and directionally controlled perturbations. Unlike indiscriminate model weights perturbations, SRMU employs a structured misdirection vector with an activation importance map. The goal is to allow SRMU selectively suppresses harmful representations while preserving the utility on benign ones. Experiments are conducted on the widely used WMDP benchmark across low- and high-entanglement configurations. Empirical results reveal that SRMU delivers state-of-the-art unlearning performance with minimal utility losses, and remains effective under 20-30\% overlap where existing baselines collapse. SRMU provides a robust foundation for safety-driven model governance, privacy compliance, and controlled knowledge removal in the emerging LLM-based applications. We release the replication package at \url{https://figshare.com/s/d5931192a8824de26aff}.

%% file: section/introduction.tex
The rapid evolution of Large Language Models (LLMs), such as GPT and Gemini, have revealed a paradigm shift in natural language processing, driving widespread deployment across different application domains. 
However, the mechanisms that underpin their success, such as deep memorization and generative capabilities, simultaneously introduce substantial security and privacy risks \cite{liu2024jailbreak}. Due to the scale and limited controllability of their training corpora, LLMs inevitably encode and retain sensitive, private, copyright-protected, even harmful information \cite{vulchi2024exploring}. Consequently, enabling a model to selectively remove the influence of specific training data without retraining from scratch, a process known as Machine Unlearning (MU), has emerged as a critical challenge for ensuring security and privacy in modern AI systems \cite{liu2024survey,nguyen2025survey,shaik2024exploring}.

Approximate MU methods (shown in Table~\ref{tab:mu_comparison}), including gradient ascent based unlearning \cite{jang2023knowledge,yao2024machine}, distillation based scrubbing \cite{dong2025undial,wang2025balancing}, and perturbation driven approaches \cite{10.1609/aaai.v39i22.34544,pmlr-v235-li24bc}, aim to revoke the effects of targeted training data without retraining \cite{geng2025comprehensive,yan2022arcane}. Among these, a group of MU methods based on controlling model representations has emerged due to its computational efficiency. For instance, Representation Misdirection for Unlearning (RMU) removes harmful knowledge by injecting global random perturbations into latent model representations \cite{pmlr-v235-li24bc}. Yet, they share a fundamental weakness, which assumes a clean separation between forget and retain samples in the model representation space. 

In practice, LLMs exhibit high knowledge entanglement, where harmful and benign concepts sharing critical feature dimensions. It results in substantial overlap between forget and retain representations, further induces limitations for existing methods. Methods such as Gradient Ascent (GA) \cite{yao2024machine} and Negative Preference Optimization (NPO) \cite{zhang2024negative} attempt to increase the loss on forget samples while decreasing it on retain samples. Under high entanglement, shared tokens appear in both forget and retain samples, yielding conflicting gradient updates. Consequently, it leads to severe deterioration in general capability performance where harmful output cannot be suppressed. Similarly, representation level approaches, including RMU \cite{pmlr-v235-li24bc} and Adaptive RMU \cite{10.1609/aaai.v39i22.34544}, utilize perturbation which are neither feature selective nor gradient direction aware. By shifting both harmful and benign feature representations in an indiscriminate way, even moderate overlap can trigger uncontrolled semantic drift and unstable utility. 
\begin{table}[ht]
\centering
\caption{A comparative summary of representative Machine Unlearning (MU) methods (Note: Existing approaches differ in intervention level and robustness under high entanglement, highlighting the need for feature-selective unlearning mechanisms)}

\label{tab:mu_comparison}

\begin{tabular}{lccccc}
\toprule
\textbf{Method} 
& \textbf{Intervention} 
& \textbf{Core} 
& \textbf{Feature} 
& \textbf{High-Ent.} \\
& \textbf{Level} 
&  \textbf{Mechanism}
& \textbf{Selective} 
& \textbf{Robust} \\
\midrule
GA \cite{yao2024machine} 
& Logit 
& Gradient ascent 
& \textbf{$\times$ }
& Low \\
NPO \cite{zhang2024negative} 
& Logit 
& Preference optimization 
& \textbf{$\times$ }
& Medium \\
UNDIAL \cite{dong2025undial} 
& Logit 
&\makecell[c]{Token-level \\adjusted-logit distillation}
& \textbf{$\times$ }
& Low \\
RKLU \cite{wang2025balancing} 
& Logit 
&\makecell[c]{Token-level \\selective distillation}
& \textbf{$\times$ }
& Medium \\
DEPN \cite{wu-etal-2023-depn}
& Neuron 
& Neuron editing 
& Partial 
& Low \\
\midrule
RMU \cite{pmlr-v235-li24bc} 
& Representation 
& Random perturbation 
& \textbf{$\times$ }
& Medium \\
Adaptive RMU \cite{10.1609/aaai.v39i22.34544} 
& Representation 
& Random perturbation 
& \textbf{$\times$ }
& Medium \\
\textbf{SRMU (Ours)} 
& Representation 
&\makecell[c]{Feature-selective\\Directional perturbation}
& \textbf{$\checkmark$}
& \textbf{High} \\
\bottomrule
\end{tabular}
\end{table}

In this work, we present \textbf{Selective Representation Misdirection for Unlearning (SRMU)}, a representation level unlearning framework designed to address the fundamental limitations of existing perturbation based methods. SRMU integrates a \emph{Dynamic Importance Map} to identify feature dimensions associated with target knowledge and a \emph{Directional Misdirection Vector} to enable controlled semantic displacement within the representation space. Experiments on the WMDP benchmark show that SRMU outperforms other state-of-the-art methods across different evaluated settings. Furthermore, SRMU remains effective under high entanglement settings where prior perturbation based approaches fail. The main contributions of this work are summarized as follows:
\begin{itemize}
    \item \textbf{Failure Mode Analysis of Approximate Unlearning.} We systematically investigate the reasons that existing MU methods fail under high entanglement scenarios, identifying the lack of feature selectivity as the root cause of utility degradation and unstable forgetting performance.

    \item \textbf{Feature-Selective Unlearning Framework.} We propose \emph{Selective Representation Misdirection for Unlearning (SRMU)}, a representation-level framework that isolates knowledge-related feature subspaces through importance-aware masking and directionally constrained perturbations.

    \item \textbf{Empirical Trade off Improvement.} By coupling precise activation suppression on target knowledge with limited interference on benign representations, SRMU significantly advances the balance between effective forgetting and utility preservation outperforming existing unlearning approaches.

    \item \textbf{Robustness under High Entanglement Settings.} Extensive evaluation on WMDP benchmark show that SRMU maintains robust performance even in scenarios that forget and retain knowledge are highly entangled, where state-of-the-art approximate methods fail to converge.
\end{itemize}


%% file: section/background.tex
\textbf{MU} tackles a fundamental problem in model security and privacy by enabling the selective removal of the influence of specific training data from trained models, while avoiding the substantial computational overhead associated with full retraining~\cite{kumar2025machine,qu2025frontier}. Since its formalization by Cao and Yang~\cite{7163042}, MU has evolved from a niche privacy technique into a fundamental paradigm for model maintenance, marking a shift from traditional knowledge accumulation to targeted, active forgetting~\cite{nguyen2025survey}. While early taxonomies categorized MU strategies into various types such as localized parameter modification or input-based filtering~\cite{geng2025comprehensive}, for modern Large Language Models (LLMs), the research landscape is predominantly divided into Exact Unlearning and Approximate Unlearning. The former provides strong theoretical guarantees but is computationally infeasible for large-scale models, whereas the latter encompasses representation-based approaches that emphasize computational efficiency and utility preservation, which this work builds upon.

\noindent\textbf{Unlearning Techniques and Architectures.}
Existing MU methods for large language models can be broadly categorized by the level at which the intervention is applied.
Logit-level unlearning methods modify the training objective to suppress undesirable behaviors, including gradient-ascent-based approaches and reinforcement learning formulations such as Quark~\cite{lu2022quark}. These methods increase the loss on forget samples while preserving performance on normal data, and are most effective when forget and retain signals induce sufficiently distinct gradient directions. Neuron-level approaches instead assume that specific knowledge is localized within a small subset of neurons. Representative methods such as DEPN~\cite{wu-etal-2023-depn} identify and edit so-called privacy neurons to remove memorized information, offering strong locality and interpretability under this localization assumption. In addition, auxiliary-model-based strategies, such as task arithmetic~\cite{ilharco2022editing}, manipulate model behavior through vector operations in weight space without explicitly isolating knowledge representations.

\noindent\textbf{Representation-level unlearning} has emerged as a particularly efficient and scalable paradigm.
Rather than modifying output distributions or individual neurons, these methods intervene on intermediate activations to induce forgetting with minimal parameter updates.
Representation Misdirection for Unlearning (RMU)~\cite{pmlr-v235-li24bc} exemplifies this line of work by applying controlled perturbations to hidden activations within a localized window of MLP layers.
Concretely, RMU updates the target layer activations as
\begin{equation}
M_{\text{updated}}(t) = M_{\text{frozen}}(t) + c \cdot \mathbf{u},
\label{eq:rmu_perturb}
\end{equation}
where $M_{\text{frozen}}(t)$ denotes the frozen reference activation, $\mathbf{u}$ is a randomly sampled perturbation direction, and $c$ controls the perturbation magnitude.
By restricting parameter updates to a single layer while perturbing a short layer window, RMU achieves computational efficiency and scalability.
Subsequent analysis by Dang et al.~\cite{10.1609/aaai.v39i22.34544} identified optimization instability in deeper layers and proposed Adaptive RMU, which dynamically rescales $c$ based on activation norms to improve training stability. Although existing representation-level approaches are efficient, they rely on unstructured, feature-agnostic perturbations, highlighting the need for selective and feature-aware modulation to control knowledge encoded in overlapping representation subspaces.

\noindent\textbf{Benchmark and Evaluation Metrics.}
Evaluating the effectiveness of unlearning in LLMs requires rigorous benchmarks tailored to specific unlearning scenarios. Existing datasets typically focus on distinct application domains. TOFU \cite{mainitofu} targets the removal of synthetic identity information for privacy protection, while WHP (Who’s Harry Potter) \cite{eldan2023s} addresses the erasure of copyrighted content within specific semantic themes. In addition, MUSE \cite{shi2024muse} and RWKU \cite{jin2024rwku} provide evaluation frameworks for unlearning in real world news and encyclopedic contexts. For safety critical hazardous knowledge, the Weapons of Mass Destruction Proxy (WMDP) Benchmark \cite{pmlr-v235-li24bc} serves as a specialized standard. Unlike privacy or copyright focused datasets, WMDP is designed to assess a model’s ability to forget domain specific factual knowledge in high risk areas such as biology (\textit{WMDP Bio}) and cyber security (\textit{WMDP Cyber}). Due to its emphasis on disentangling hazardous knowledge from general reasoning capabilities, WMDP is selected as the primary testbed for our experiments to evaluate the \textit{semantic selectivity} of the proposed SRMU framework.

%% file: section/method.tex
Our proposed \emph{Selective Representation Misdirection for Unlearning} (SRMU) is shown in Figure~\ref{fig:selectivermu}. Instead of using unstructured perturbations, we present a novel solution for selective and feature aware modulation.
SRMU aims to remove the influence of a \emph{forget dataset} $\mathcal{D}_{\text{forget}}$ while preserving model behavior on a \emph{retain dataset} $\mathcal{D}_{\text{retain}}$ through targeted intervention on intermediate representations.
We define $M_{\text{frozen}}$ as the pretrained model and $H_{\theta}(x)$ as the activation output at a designated multilayer perceptron (MLP) layer.
During unlearning process, activations from the MLP layer are used to compute forgetting and retention losses. Only the parameters of the target layer are updated to minimise our unlearning objective, while all other layers remain frozen.

\subsection{Feature Aware Representation Misdirection}

The SRMU framework extends RMU via three key components, as explained in the following subsections.
First, perturbations are modeled in a feature aware manner, ensuring that interventions are guided by the semantic relevance of internal representations.
Second, dynamic importance maps are constructed to identify feature dimensions most strongly associated with the target knowledge.
Third, a selective optimization mechanism is applied to constrain updates to high importance regions, thereby minimizing interference with unrelated capabilities.

\subsection{Dynamic Importance Map Construction}\label{sec:importance_modeling}

The importance weights $I$ are not fixed parameters but are dynamically computed based on the input samples.
To quantify feature level relevance, SRMU extracts intermediate feature representations and constructs a \textbf{Dynamic Importance Map} by comparing activation patterns between the forget set ($v_f$) and the retain set ($v_r$). Here, $v_f$ and $v_r$ denote the averaged hidden activations of the target layer
over mini-batches sampled from $\mathcal{D}_{\text{forget}}$ and $\mathcal{D}_{\text{retain}}$, respectively:
\begin{equation}
I = \phi(v_f, v_r),
\label{eq:mapcal}
\end{equation}
where $\phi(\cdot)$ denotes an importance fusion function that measures the contribution of each feature dimension to the forgetting objective.
We consider three instantiations of the importance function $\phi(\cdot)$, each capturing a distinct notion of feature relevance between the forget and retain sets:
\begin{itemize}
    \item \textbf{SRMU Ratio based:}
    \begin{equation*}
    I_{\text{ratio}} = \log\left(1 + \frac{v_f}{v_r + \epsilon}\right)
    \end{equation*}
    This formulation emphasizes feature dimensions where forgetting activations dominate retention signals.
    Logarithmic scaling is applied to stabilize extreme ratios, prioritizing features with strong relative forgetting strength.

    \item \textbf{SRMU Difference based:}
    \begin{equation*}
    I_{\text{diff}} = \text{ReLU}(v_f - \lambda v_r)
    \end{equation*}
    This strategy isolates features that are highly activated for the forget set but suppressed for the retain set, resulting in a sparse and selective importance map.

    \item \textbf{SRMU Product based:}
    \begin{equation*}
    I_{\text{prod}} = \frac{v_f \odot v_r}{\text{mean}(v_f) \cdot \text{mean}(v_r) + \epsilon}
    \end{equation*}
    This formulation highlights dimensions that are simultaneously activated by both sets, explicitly identifying entangled features that require cautious and controlled perturbation.
\end{itemize}

The resulting importance map reflects the extent to which each representation dimension contributes to the target forgetting process.
To ensure stability and comparability when integrating with the Directional Misdirection Vector $\mathbf{V}$, the importance map $I$ is normalized across all feature dimensions:
\begin{equation}
I_{\text{norm}} = \frac{I}{\max(I) + \epsilon_{\text{norm}}},
\label{eq:normalization}
\end{equation}
where $\epsilon_{\text{norm}} = 10^{-8}$ prevents division by zero.
This normalization scales importance values to the range $[0,1]$, enabling consistent control via the coefficient $c_{\text{map}}$ in subsequent optimization.

\subsection{Directional Misdirection Vector ($\mathbf{V}$) Generation}\label{sec:vector}

The \textbf{Directional Misdirection Vector} ($\mathbf{V}$) is introduced to address the non selective and direction agnostic nature of the original RMU approach.
In RMU, the perturbation direction $u$ is sampled as a purely random unit vector, providing no explicit semantic control.
To overcome this limitation, SRMU redefines the perturbation direction as the Directional Misdirection Vector $\mathbf{V}$.
Specifically, $\mathbf{V}$ is constructed as a discrete, high dimensional vector:
\[
\mathbf{V} \in \{-1, +1\}^d,
\]
where each element is sampled independently.
This design enables controlled semantic deviation by enforcing a consistent polarity for each feature dimension, thereby defining a structured trajectory away from the original knowledge encoding. Compared to continuous random directions, this discrete formulation provides
stable and interpretable directional control, as validated in our ablation study.
\subsection{Final Loss Function of SRMU}

SRMU jointly enforces targeted forgetting and representation preservation through a two term objective.
For a forget sample $x_f$, we define a \emph{Directional Misdirection Target} that specifies the desired activation state along selected feature dimensions:
\begin{equation}
T_{\text{misdir}}(x_f) = c_{\text{map}} \cdot \mathbf{V} \odot I_{\text{norm}}(x_f),
\label{eq:misdirection_target_final}
\end{equation}
where $I_{\text{norm}}(x_f)$ denotes the normalized importance map and $\mathbf{V}$ is the directional misdirection vector.
This target defines a feature wise displacement that the updated representation is encouraged to follow.

The final optimization objective is given by:
\begin{equation}
\begin{aligned}
\mathcal{L}_{\text{SRMU}}
= \;&
\mathbb{E}_{x_f \sim \mathcal{D}_{\text{forget}}}
\left[
\| H_\theta(x_f) - T_{\text{misdir}}(x_f) \|_2^2
\right] \\
&+
\alpha \,
\mathbb{E}_{x_r \sim \mathcal{D}_{\text{retain}}}
\left[
\| H_\theta(x_r) - H_{\theta_0}(x_r) \|_2^2
\right].
\end{aligned}
\end{equation}
The first term enforces targeted displacement along forget-relevant dimensions,
while the second term anchors retain representations to their original states.
Here, $H_{\theta_0}$ denotes the frozen pretrained model and $\alpha$ controls the trade off between forgetting effectiveness and representation preservation.
Minimizing $\mathcal{L}_{\text{SRMU}}$ with respect to the target MLP layer completes the SRMU update.

\subsection{Pseudocode and Complexity}
\begin{algorithm}[H]
\caption{Selective Representation Misdirection for Unlearning (SRMU)}
\DontPrintSemicolon
\KwIn{
Updated model $M_{updated}$, frozen model $M_{frozen}$,
forget set $\mathcal{D}_{forget}$, retain set $\mathcal{D}_{retain}$,
scaling coefficient $c_{map}$, retain weight $\alpha$
}
\KwOut{Unlearned model $M_{updated}$}

Compute importance map $I_{norm}$ from $\mathcal{D}_{forget}$ and $\mathcal{D}_{retain}$\;
Sample directional vector $\mathbf{V} \in \{-1,+1\}^d$\;

\For{$x_f \sim \mathcal{D}_{forget},\; x_r \sim \mathcal{D}_{retain}$}{
    Set misdirection target $T_{misdir} = c_{map} \cdot (\mathbf{V} \odot I_{norm})$\;
    Set $\mathcal{L}_{forget} = \| M_{updated}(x_f) - T_{misdir} \|_2^2$\;
    Set $\mathcal{L}_{retain} = \| M_{updated}(x_r) - M_{frozen}(x_r) \|_2^2$\;
    Update $M_{updated}$ using $\mathcal{L} = \mathcal{L}_{forget} + \alpha \mathcal{L}_{retain}$\;
}

\Return{$M_{updated}$}
\end{algorithm}

\subsubsection{Complexity}
SRMU incurs a one time overhead to construct the Dynamic Importance Map $\mathbf{I}_{\text{norm}}$, which is computed prior to optimization.
This cost is dominated by forward passes through the target layer $l$ over $\mathcal{D}_{\text{map}} = \mathcal{D}_{\text{forget}} \cup \mathcal{D}_{\text{retain}}$, yielding a complexity of
\[
\mathcal{O}(|\mathcal{D}_{\text{map}}| \cdot \mathcal{T}_{\text{forward}}(l)).
\]
The subsequent importance fusion and normalization steps involve only element wise operations and incur negligible additional cost.

During unlearning, SRMU performs $T$ optimization steps, each updating only the designated target layers.
The per step cost is therefore
\[
\mathcal{O}(\mathcal{T}_{\text{forward}}(\mathcal{M}_{\text{target}}) + \mathcal{T}_{\text{backward}}(\mathcal{M}_{\text{target}})),
\]
which is substantially lower than full model fine tuning.
Considering these factors, SRMU preserves the computational efficiency and scalability of perturbation based machine unlearning methods like RMU.

%% file: section/new_experiment.tex
\subsection{Experimental Setup}
All methods are evaluated on the Zephyr 7B model, following prior work on RMU and Adaptive RMU to ensure fair comparison. Zephyr 7B provides stable access to intermediate MLP representations, which is required for representation level unlearning.\par
\noindent\textbf{Datasets.}
For each WMDP domain (Biology and Cybersecurity), we construct a forget set $\mathcal{D}_{\text{forget}}$ and evaluate unlearning under two retain regimes:
(i) \textbf{Low entanglement}, where the retain set is drawn from WikiText; and
(ii) \textbf{High entanglement}, where the retain set is drawn from the same domain as $\mathcal{D}_{\text{forget}}$.

In the high entanglement setting, prior analyses \cite{10.1609/aaai.v39i22.34544} report substantial overlap between forget and retain sets. Specifically, the Biology retain set exhibits 20.8\% unigram and 5.5\% bigram overlap, while the Cybersecurity retain set shows 27.5\% unigram and 12.3\% bigram overlap. Such overlap indicates that harmful and benign samples activate highly shared representation regions, making selective unlearning more challenging. These two regimes enable evaluation of SRMU under both controlled and adversarial unlearning conditions.

\noindent\textbf{Evaluation Metrics.}
We evaluate the unlearning utility trade off using two standardized benchmarks:
\begin{itemize}
    \item \textbf{WMDP Accuracy ($\downarrow$)}: measures the remaining ability to answer hazardous Biology and Cybersecurity questions, where lower accuracy indicates stronger forgetting.
    \item \textbf{MMLU Accuracy ($\uparrow$)}: measures general language understanding across 57 subjects, where higher accuracy reflects better utility preservation.
\end{itemize}
Full baseline comparisons (LLMU, SCRUB, SSD) are conducted under the low entanglement regime, while high entanglement experiments focus on perturbation based methods (RMU, Adaptive RMU) and SRMU.

\noindent\textbf{Training.}
To ensure a fair comparison, SRMU adopts the same optimization configuration utilized in prior studies \cite{10.1609/aaai.v39i22.34544,pmlr-v235-li24bc}.
We use AdamW with a learning rate of $5\times10^{-5}$, batch size 4, and $T=150$ unlearning steps.
The retain loss weight is set to $\alpha=1200$, and the RMU perturbation magnitude is fixed at $c=7.5$, following prior work.
For RMU and Adaptive RMU, perturbation budget is selected via grid search over $[1,170]$ with a step size of 10.
SRMU adopts the same procedure to select $c_{\text{map}}$, ensuring comparable perturbation scales across methods.
Unless otherwise stated, the sequence length is set to 512 for Biology and 768 for Cybersecurity.

\begin{table}[!h]
\centering
\caption{Comparative performance of SRMU against SOTA baselines on Forgetting (WMDP) and Retention (MMLU).}
\label{tab:overall_results}

\begin{tabular}{lcccc}
\toprule
\textbf{Method} 
& \textbf{MMLU ($\uparrow$)} 
& \makecell{\textbf{WMDP-}\\\textbf{Bio ($\downarrow$)}} 
& \makecell{\textbf{WMDP-}\\\textbf{Cyber ($\downarrow$)}} 
& \makecell{\textbf{WMDP}\\\textbf{Avg ($\downarrow$)}} \\
\midrule
Original (Zephyr-7B) & 58.5 & 64.7 & 44.8 & 54.7 \\
LLMU (Yao et al., 2024) & 44.7 & 59.5 & 39.5 & 49.5 \\
SCRUB (Kurmanji et al., 2023) & 51.2 & 43.8 & 39.3 & 41.6 \\
SSD (Foster et al., 2024) & 40.7 & 50.2 & 35.0 & 42.6 \\
\midrule
RMU (Li et al., 2024) & 56.9 & 28.8 & 28.0 & 28.4 \\
Adaptive RMU (Dang et al., 2025) & 55.0 & \textbf{25.3} & 26.7 & \textbf{26.0} \\
\textbf{SRMU (Our)} & \textbf{57.1} & 28.5 & \textbf{25.8} & 27.2 \\
\bottomrule
\end{tabular}
\end{table}
\subsection{Main Results}
\textbf{Reproduction Note.}
Results for RMU and Adaptive RMU are reproduced on the Zephyr 7B model following the hyperparameter search strategy described in the Experimental Setup, and the best performing results are reported in Table~\ref{tab:overall_results}.
Results for LLMU, SCRUB, and SSD are directly cited from prior work \cite{10.1609/aaai.v39i22.34544}.

In the low entanglement setting, SRMU achieves the best overall unlearning utility trade off.
It reduces the WMDP average from 54.7\% to 27.2\%, matching the forgetting strength of Adaptive RMU while preserving higher utility (57.1\% MMLU versus 55.0\%).
Compared with RMU, SRMU improves both forgetting (27.2\% versus 28.4\%) and retention (57.1\% versus 56.9\%), indicating that importance aware perturbation mitigates unnecessary drift caused by direction agnostic misdirection.
Overall, SRMU defines the Pareto frontier in the low entanglement regime and provides a stable reference for evaluation under more challenging conditions.

\subsection{Results under the High Entanglement Regime}

Table~\ref{tab:hard_regime} reports results when the retain set is drawn from the same domain as the forget set, resulting in strong representation entanglement. Under this regime, RMU and Adaptive RMU exhibit limited forgetting at comparable MMLU levels, with WMDP accuracy remaining high, indicating substantial retention of harmful knowledge.

\begin{table}[!h]
\centering
\caption{Performance of RMU, Adaptive RMU, and SRMU under the High-Entanglement Retain Regime.}
\label{tab:hard_regime}

\begin{tabular}{lcccc}
\toprule
\textbf{Method} 
& \textbf{MMLU ($\uparrow$)} 
& \makecell{\textbf{WMDP-}\\\textbf{Bio ($\downarrow$)}} 
& \makecell{\textbf{WMDP-}\\\textbf{Cyber ($\downarrow$)}} 
& \makecell{\textbf{WMDP}\\\textbf{Avg ($\downarrow$)}} \\
\midrule
Original (Zephyr-7B) & 58.5 & 64.7 & 44.8 & 54.7 \\
\midrule
RMU (Li et al., 2024) & 51.9 & 48.5 & 41.1 & 44.8 \\
Adaptive RMU (Dang et al., 2025) & 51.15 & 49.33 & 37.74 & 43.54 \\
\textbf{SRMU (Our)} & \textbf{52.5} & \textbf{38.26} & \textbf{37.14} & \textbf{37.7} \\
\bottomrule
\end{tabular}
\end{table}

Adaptive RMU reduces the WMDP average from 54.7 to 43.54 at an MMLU of 51.15, while RMU achieves an even smaller reduction at similar utility.
Extensive hyperparameter sweeps confirm that this limitation is not due to suboptimal tuning.

In contrast, SRMU achieves stronger forgetting under matched retention conditions. At a comparable MMLU level (52.5), SRMU further reduces the WMDP average to 37.7 while preserving similar general capability.
These results indicate that the failure of prior perturbation based methods in the high entanglement regime is structural rather than parametric.
By selectively localizing and perturbing forget relevant features, SRMU enables effective unlearning within shared representation subspaces.

\subsection{Ablation Study}

All ablation studies are conducted under the low-entanglement
(Wikitext retain) setting, where forget and retain distributions
are weakly overlapping.
\\
\par
\begin{figure}[!h]
    \centering
    \includegraphics[width=\linewidth]{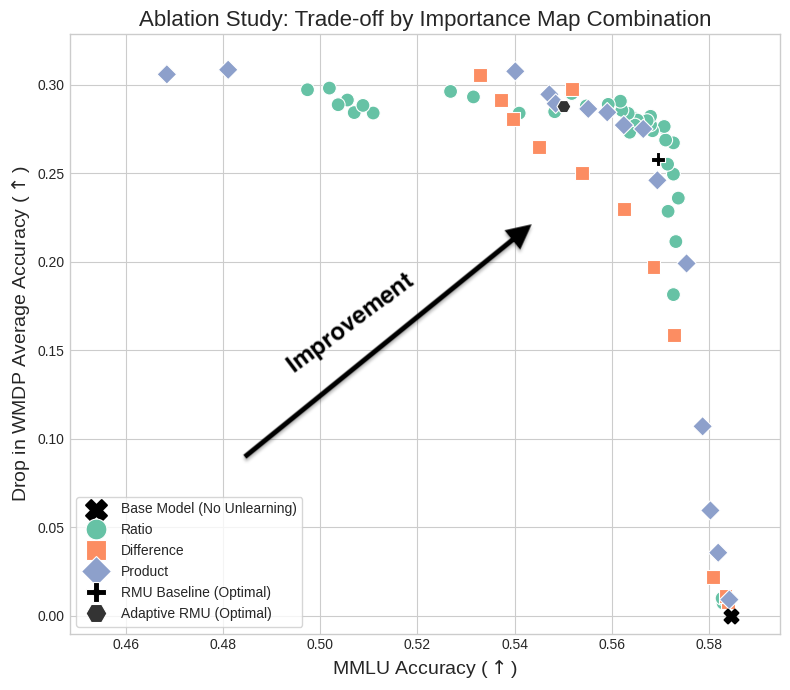}
    \caption{Trade-off Comparison of Importance Map Strategies:
    WMDP Drop vs.~MMLU Accuracy.}
    \label{fig:combinationablation}
\end{figure}

 
\noindent
\textbf{Effect of Importance Map Design:} We compare three importance map constructions for $\mathbf{I}_{\text{norm}}$: Ratio, Difference, and Product.
Figure~\ref{fig:combinationablation} shows the trade-off between WMDP reduction and MMLU accuracy.

The \textbf{Ratio-based} formulation consistently defines the Pareto frontier, achieving higher utility at the same forgetting level than both alternative SRMU variants and RMU-based baselines.
The \textbf{Product-based} formulation performs competitively in intermediate regions but does not reach the frontier, while the \textbf{Difference-based} formulation underperforms due to overly sparse importance assignments.

Overall, these results indicate that appropriate importance weighting is critical, with the Ratio-based strategy providing the best forgetting–utility balance.
\\
\par 
\noindent
\textbf{Effect of Directional Misdirection:} Table~\ref{tab:ablation_results} evaluates the necessity of directional misdirection
($\mathbf{V}$) and importance normalization ($\mathbf{I}_{\text{norm}}$).
When $\mathbf{T}=0$, performance remains close to the base model
(MMLU $0.5835$, WMDP Avg $0.5394$), indicating negligible forgetting.
Fixed direction variants achieve strong forgetting
(WMDP Avg $\approx 0.25$) but suffer severe utility degradation
(MMLU $<0.30$), while removing $\mathbf{I}_{\text{norm}}$ also degrades
utility (MMLU $0.5165$).
In contrast, combining adaptive directional perturbation with importance
normalization yields effective forgetting while preserving utility,
confirming that both components are necessary for SRMU.

\begin{table}[H]
\centering
\caption{Ablation Study of SRMU Components under the Low-Entanglement Setting.}
\label{tab:ablation_results}

\begin{tabular}{lcc}
\toprule
\textbf{Method / Variant} 
& \textbf{MMLU ($\uparrow$)} 
& \textbf{WMDP Avg ($\downarrow$)} \\
\midrule
Base Model & 0.5845 & 0.5475 \\
RMU Baseline & 0.5694 & 0.2896 \\
\textbf{SRMU (Full)} & \textbf{0.571} & \textbf{0.2717} \\
\midrule
\multicolumn{3}{l}{\textbf{Mechanistic Ablation}} \\
SRMU (w/o $\mathbf{V}$ and $\mathbf{I}_{\text{norm}}$; $\mathbf{T}=0$) & 0.5835 & 0.5394 \\
SRMU (w/o $\mathbf{I}_{\text{norm}}$; uniform perturbation) & 0.5165 & 0.2503 \\
SRMU (fixed $+1$ direction) & 0.2876 & 0.2430 \\
SRMU (fixed $-1$ direction) & 0.2483 & 0.2556 \\
SRMU (random direction in $[0,1)$) & 0.5672 & 0.2755 \\
\bottomrule
\end{tabular}
\end{table}

%% file: section/conclusion.tex
In this work, we presented Selective Representation Misdirection for Unlearning (SRMU), a representation level unlearning framework that addresses the limitations of random and non selective perturbation. SRMU integrates a Dynamic Importance Map with a Directional Misdirection Vector to enable feature selective and controlled modification of internal representations. Experiments on WMDP demonstrated that SRMU achieves a superior balance between forgetting effectiveness and utility preservation.
SRMU matches or outperforms existing perturbation based methods under standard settings and remains effective in high entanglement regimes where prior approaches degrade. Ablation results further demonstrated that both importance aware feature selection and directional misdirection are necessary for stable and selective unlearning. Overall, these results indicate that effective machine unlearning in large language models requires structured, feature aware intervention rather than unstructured perturbation. By enabling targeted representation editing under realistic entanglement conditions, SRMU provides a practical and interpretable framework for compliance driven knowledge removal.